\documentclass[10pt,twocolumn,letterpaper]{article}

\usepackage{cvpr}
\usepackage{times}
\usepackage{epsfig}
\usepackage{graphicx}
\usepackage{amsmath}
\usepackage{amssymb}
\usepackage{tikz}

\usepackage[breaklinks=true,bookmarks=false]{hyperref}

\cvprfinalcopy 

\def\cvprPaperID{****} 

\begin{document}

\title{FusionNet: 3D Object Classification Using Multiple Data Representations}

\author{Vishakh Hegde\\
Stanford and Matroid\\
{\tt\small vishakh@matroid.com}
\and
Reza Zadeh\\
Stanford and Matroid \\
{\tt\small reza@matroid.com}
}

\maketitle

\begin{abstract}
High-quality 3D object recognition is an important component of many vision and robotics systems. We tackle the object recognition problem using two data representations, to achieve leading results on the Princeton ModelNet challenge. The two representations:
\begin{itemize}
\item Volumetric representation: the 3D object is discretized spatially as binary voxels - $1$ if the voxel is occupied and $0$ otherwise.
\item Pixel representation: the 3D object is represented as a set of projected 2D pixel images.
\end{itemize}
Current leading submissions to the ModelNet Challenge use Convolutional Neural Networks (CNNs) on pixel representations. However, we diverge from this trend and additionally, use Volumetric CNNs to bridge the gap between the efficiency of the above two representations. We combine both representations and exploit them to learn new features, which yield a significantly better classifier than using either of the representations in isolation. To do this, we introduce new Volumetric CNN (V-CNN) architectures.
\end{abstract}

\section{Introduction}

Three dimensional model classification is an important problem, with applications including self driving cars and augmented reality, among others. 3D content creation has been picking up pace in the recent past and the amount of information in the form of 3D models becoming publicly available is steadily increasing. This is good news for methods based on Convolutional Neural Networks (CNNs) whose performance currently rely heavily on the availability of large amount of data. A lot of methods based on CNNs have been proposed recently, for the purposes of image and object classification. They achieve significantly better results compared to using standard machine learning tools on top of hand crafted features.

The real world is three dimensional. Compared to 2D visual information, currently very little 3D data is being generated. However, with the advent of technology such as augmented reality and self driving cars, 3D data is becoming more and more ubiquitous as a form of information content. Not in the distant future, applications of 3D data of objects will use classification algorithms crafted for 3D content in order to achieve certain target tasks (like robot navigation in a cluttered environment or smart user interfaces based on augmented reality). With advances in 3D scanning technology and use of augmented Reality in daily life, it is important to come up with algorithms and methods to classify 3D models accurately. 

In this paper, we present our work on classifying 3D CAD models bench-marked on a standard 3D CAD model dataset called the Princeton ModelNet \cite{3DShapenets}. With increase in the availability of 3D data, a lot of deep learning algorithms are being designed for classification and detection purposes \cite{zadeh2013dimension,zadeh2015matrix}. However, the task of designing deep networks suitable for 3D data is not trivial as one might assume. For example, most information in RGB images is encoded as pixels and pixel intensity distribution for each color channel. However, information of 3D models reside on the surface and relative orientation of meshes which define the model surface. Therefore, features useful for 2D image classification might not necessarily be sufficient for 3D model classification. CAD models are sometimes devoid of color (as is the case for the dataset we use) and therefore our models must be able to pick up on other features which define the 3D model.

Volumetric representation in the form of binary voxels was shown by \cite{3DShapenets}, to be useful for classification and retrieval of 3D models. They train their network generatively. \cite{voxnet} introduced Voxnet, a 3D CNN for 3D point cloud data and voxelized models, which performed significantly better than \cite{3DShapenets}. In ~\cite{deeppano}, the authors suggest a new robust representation of 3D data by way of a cylindrical panoramic projection that is learned using a CNN. The authors tested their panoramic representation on ModelNet datasets and outperformed typical methods when they published their work. There was a significant jump in classification and retrieval performance by simply using 2D projections of the 3D model and using networks pre-trained on ImageNet \cite{imagenet_cvpr09} for classification as shown by \cite{MVCNN}. A part of this significant jump in performance is due to highly efficient and data independent features learned in the pre-training stage, which generalize well to other 2D images. However, this difference in classification performance between CNNs that use volumetric input and those that use 2D pixel input was partly bridged in a concurrent paper on Volumetric CNNs \cite{qi2016volumetric}.

In our work, we introduce two new CNNs for volumetric data with significantly less number of parameters compared to standard CNNs used on 2D RGB images such as AlexNet. One of these networks is inspired by the inception module used in GoogLeNet \cite{DBLP:journals/corr/SzegedyLJSRAEVR14}. These volumetric CNNs complement the strengths of our MV-CNN which uses AlexNet pre-trained on ImageNet. In particular, we combine these networks in a way which improves on the state of the art classification accuracy.

\section{Related Work}
\subsubsection*{Shape descriptors}
A large body of literature in the computer vision and graphics research community has been devoted to designing shape descriptors for 3D objects. Depending on data representations used to describe these 3D models, there has been work on shape descriptors for voxel representations and point cloud representation, among many others. In the past, shapes have been represented as histograms or bag of features models which were constructed using surface normals and surface curvatures \cite{horn1983extended}. Other shape descriptors include the Light Field Descriptor \cite{Chen03onvisual}, Heat kernel signatures \cite{Bronstein:2011:SGG:1899404.1899405} \cite{Kokkinos12intrinsicshape} and SPH \cite{Kazhdan03harmonic}. Classification of 3D objects has been proposed using hand-crafted features along with a machine learning classifier in \cite{lai20093d}, ~\cite{teichman2011towards} and ~\cite{behley2012performance}. However, more recently, the focus of research has also included finding better ways to represent 3D data. In a way, better representation has enabled better classification. The creators of the Princeton ModelNet dataset have proposed a volumetric representation of the 3D model and a 3D Volumetric CNN to classify them ~\cite{3DShapenets}. 

With recent improvements in the field of deep learning, Convolutional Neural Networks (CNN) have been widely and successfully used on 2D RGB images for a variety of tasks in computer vision, such as image classification \cite{DBLP:journals/corr/DonahueJVHZTD13} ~\cite{Simonyan14c}, object detection, semantic segmentation  \cite{DBLP:journals/corr/GirshickDDM13} ~\cite{DBLP:journals/corr/RedmonDGF15} and scene recognition \cite{farabet-pami-13}.

More recently, they have also been used to perform classification and retrieval of 3D CAD models ~\cite{3DShapenets} ~\cite{voxnet} ~\cite{deeppano} ~\cite{MVCNN} ~\cite{DBLP:journals/corr/JohnsLD16} ~\cite{qi2016volumetric}. CNNs not only allow for end to end training, it is also an automated feature learning method. The features learned through CNNs generalize well to other datasets, sometimes containing very different category of images. In particular, the distributed representation of basic features in different layers and different neurons means that there are a huge number of ways to aggregate this information in order to accomplish a task like classification or retrieval. It is also known that the features learned by training from a large 2D RGB image dataset like ImageNet generalize well, even to images not belonging to the original set of target classes. This is in contrast to handcrafted features which do not necessarily generalize well to other domains or category of images.
\section{Methods}
\label{sec:methods_section}
Most state of the art 3D object classification algorithms are based on using Convolutional Neural Networks (CNNs) to discriminate between target classes \cite{voxnet} ~\cite{deeppano} ~\cite{MVCNN} ~\cite{DBLP:journals/corr/JohnsLD16} ~\cite{qi2016volumetric}. Generative models on voxels have also been used for this problem \cite{3DShapenets}. Most approaches used to solve the 3D object classification problem involves two steps namely, deciding a data representation to be used for the 3D object and training a CNN on that representation of the object. Most state of the art methods either use voxels \cite{3DShapenets} ~\cite{voxnet} ~\cite{qi2016volumetric} or a set of multiple 2D projections of the polygon mesh from several camera positions \cite{deeppano} ~\cite{MVCNN} ~\cite{DBLP:journals/corr/JohnsLD16}.

We propose a method based on Convolutional Neural Networks, which uses both voxel and pixel representations for training relatively weak classifiers. We combine them to obtain a relatively stronger, more discriminative model. For pixel data, we use ideas from Multi View CNN (MV-CNN) proposed by ~\cite{MVCNN} which use multiple projected views of the same 3D model and aggregate them in a way which improves the performance over a single projection. For voxel data, we use our own neural networks on multiple orientations of all objects in the training set to learn features which are partly complementary to those learned for pixel data. In particular, one of our Volumetric CNN has about 3.5 million parameters (compared to AlexNet, which has 60 million parameters \cite{NIPS2012_4824}). Therefore, it can be used as an add on network to MV-CNN without increasing the number of parameters by too much. Finally in FusionNet, we combine multiple networks in the final fully connected layer which outputs class scores in a way which improves the classification accuracy over each of its component networks.

\section{Dataset and Accuracy Measure}
The use of CNN for 3D model classification had been limited due to unavailability of large scale training sets like ImageNet. This problem is partly solved by the introduction of ModelNet which contains a total of $662$ object classes, $127,915$ CAD models and subsets like ModelNet40 and ModelNet10, which are used to benchmark classification algorithms for 3D models. 

We make use of ModelNet40, which consists of $12311$ distinct CAD models belonging to one of $40$ classes. We use the same train-test split provided by the authors of the dataset, which have $9843$ models for training and $2468$ models for testing. We also test our methods on ModelNet10, which is a subset of ModelNet40, consisting of $4899$ CAD models. For ModelNet10, we are guaranteed that the orientations of all CAD models are aligned along their gravity axis. We use the same train-test split provided by the authors of the dataset, which have $3991$ CAD models for training and $908$ CAD models for testing.

For both ModelNet10 and ModelNet40 datasets, we report the average per-class classification accuracy, which is the accuracy measure used in most previous work.

\section{Volumetric CNN (V-CNN)} \label{section:V-CNN}
We introduce two CNNs for volumetric data which constructively combines information from multiple orientations by max-pooling across 60 orientations. This is similar to \cite{MVCNN} which aggregates multiple 2D projections of a 3D polygonal mesh using a well established CNN architecture on 2D RGB images, like VGG-M \cite{Simonyan14c}. Both our networks also use long range 3D convolutions which aggregate information across a dimension of the object. In one of the networks, we concatenate the output from kernels of various sizes, similar to the inception module used in \cite{DBLP:journals/corr/SzegedyLJSRAEVR14}. In section \ref{section:Experiments}, we compare the classification accuracy of the two models.

\begin{figure}[t] 
\begin{center}
	\includegraphics[scale=0.2]{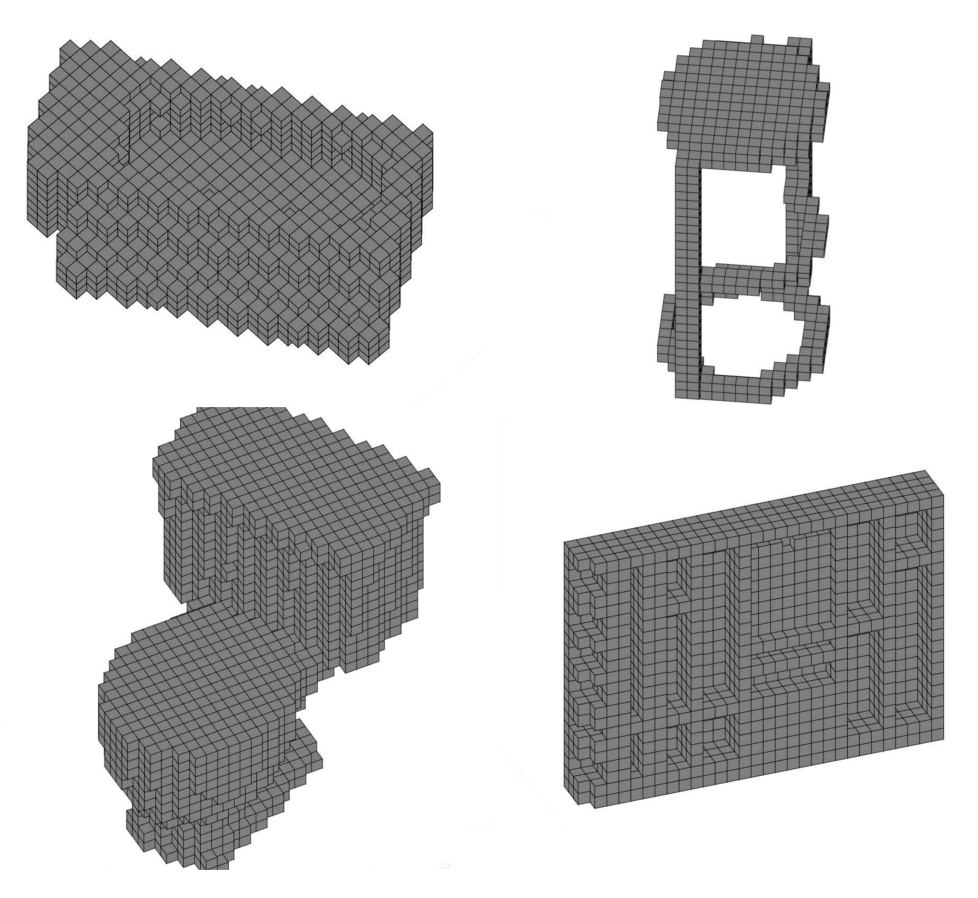}
\end{center}
   \caption{Sample voxelized version of 3D CAD models from ModelNet40. Top-left: bathtub, top-right: stool, bottom-left: toilet, bottom-right: wardrobe.}
\label{voxel}
\end{figure}

\subsection{Data Augmentation}
Unlike the presence of large scale 2D RGB image datasets like ImageNet, there are relatively fewer number of CAD models. This translates to having fewer number of voxelized input which can be fed to a Volumetric CNN. This will result in features that are not as efficient as those learned for 2D images. Therefore, it is necessary to augment training data with multiple rotations of voxelized models. We input multiple azimuth and polar rotations of the model to the Volumetric CNN, apart from another non-standard data augmentation method explained in \ref{subsub:DataAug}.

We work under the assumption that all CAD models in the dataset are aligned along the gravity axis. We produce voxels of size $30 \times 30 \times 30$ from polygon meshes after rendering them in 60 different orientations about the gravity axis, where each rendering is defined by $\theta$, the polar angle and $\phi$, the azimuth angle. To obtain these 60 orientations, we uniformly sample 60 polar angles from $[0, \pi]$ and 60 azimuth angles from $[0, 2 \pi]$ and use each pair of polar and azimuth angles for rendering voxels. 

We perform random sampling of angles to break any symmetry present along the gravity axis. Otherwise, no new information is added for objects which are symmetric about the gravity axis (for example, symmetric objects like vase and bowl). We hope that inputting multiple such random rotations will help the neural network learn rotational invariance, both in the polar angle space and in the azimuth angle space. This is especially important if the polygon mesh being tested on does not have the same polar angle as most meshes belonging to that class in the training set. This is also important for classifying models constructed from real world data like RGB-D and LiDAR data, where rotational invariance is necessary for good performance.

\subsection{V-CNN I}
\begin{figure*}[t] \label{figure:VCNNI}
\begin{center}
	\includegraphics[width=\textwidth]{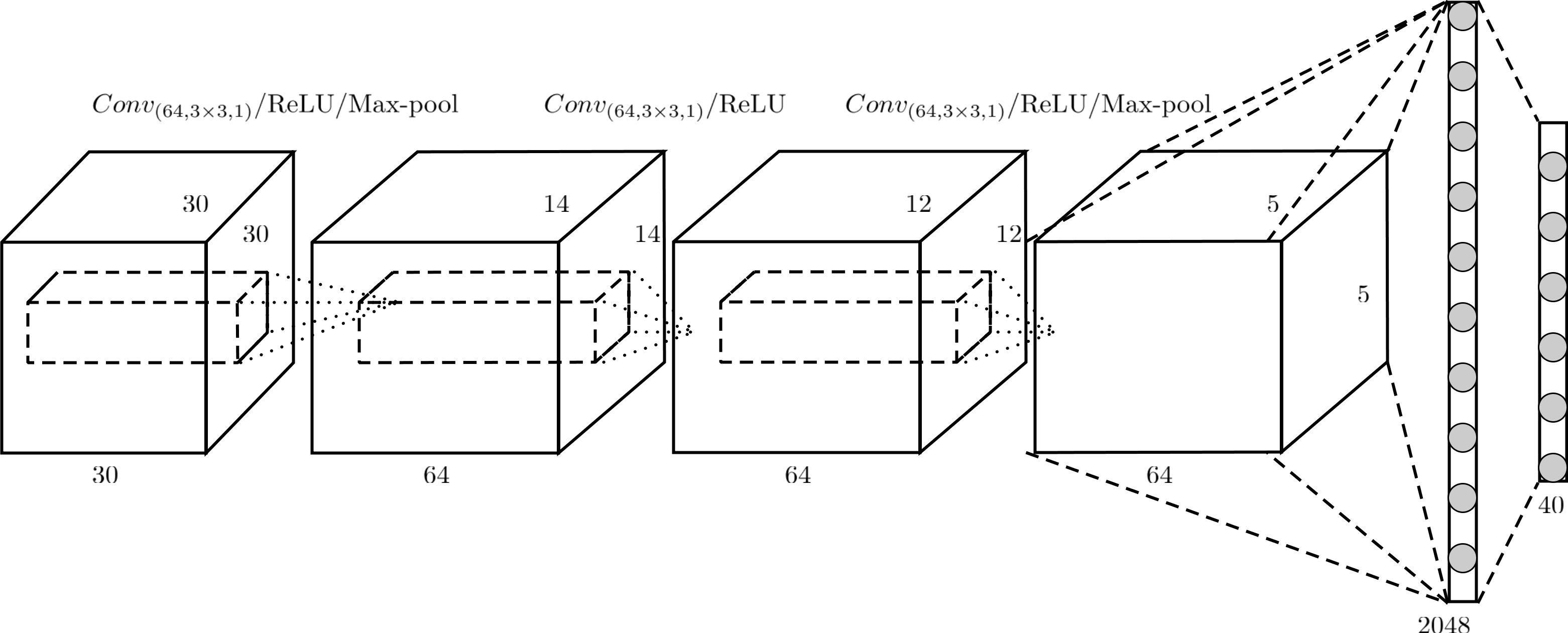}
\end{center}
   \caption{Network architecture of V-CNN I. It has three 2D convolution layers, all with kernels of size $3 \times 3$ and two fully connected layers. The first convolutional layer assumes the number of channels to be $30$. A Softmax Loss function is used while training the network.}
\label{voxel}
\end{figure*}

This Volumetric CNN consists of three 3D convolution layers and two fully connected layers, where the final fully connected layer is used as a classifier as depicted in figure \ref{figure:VCNNI}. The kernels used in the convolution layer find correlations along the full depth of the object. When trained on different orientations for all models, the hope is to be able to learn long range spatial correlations in the object along all directions while using sparse locally connecting kernels in order to reduce computational complexity. This is difficult to achieve in 2D images where kernels only adapt to spatially local pixel distributions and depend on the kernel-size used for the convolution. 

The ReLU layer \cite{DBLP:conf/icml/NairH10} following the convolution layer introduces non-linearity in the model necessary for class discrimination. The pooling layer following ReLU ensures that neurons learning redundant information from a spatially local set of voxels do not contribute to the size of the model. The kernels used in all convolution layers in this network are of size $3 \times 3$. While CNNs for 2D image classifiation like AlexNet use kernels of size $11 \times 11$, we believe that $3 \times 3$ kernels are sufficient to capture correlations for voxelized data. The reason is that a single cross-section has a resolution of $30 \times 30$ (in comparison to a resolutions of $227 \times 227$ for images used in AlexNet \cite{NIPS2012_4824}). Dropout \cite{Srivastava:2014:DSW:2627435.2670313} is used to reduce any over-fitting. A fully connected layer with $40$ neurons is used as the classifier for ModelNet40 dataset and a fully connected layer with $10$ neurons as the classifier for ModelNet10 dataset. Details of this architecture is given in table \ref{table:VCNN I}.

\begin{table*} \label{table:VCNN I}
\begin{center}
\label{CNN_tuning}
\begin{tabular}{|c|c|c|c|c|}
\hline
Type & Filter size & Stride & Output size & Number of parameters\\
\hline\hline
Convolution& $3 \times 3$ & 1 & $64 \times 28 \times 28$ & 17344\\
ReLU & -- & -- & $64 \times 28 \times 28$ & -- \\
Max pool & $2 \times 2$ & 2 & $64 \times 14 \times 14$ &--\\
& & & & \\
Convolution& $3 \times 3$ & 1 & $64 \times 12 \times 12$ & 36928\\
ReLU & -- & -- & $64 \times 12 \times 12$ & -- \\
& & & & \\
Convolution& $3 \times 3$ & 1 & $64 \times 10 \times 10$ & 36928\\
Max pool & $2 \times 2$ & 2 & $64 \times 5 \times 5$ &--\\
Dropout & 0.5 & -- & $64 \times 5 \times 5$ & --\\
& & & & \\
Fully Connected & -- & -- & 2048 & 3,278,848\\
Fully Connected & -- & -- & 40 & 81,960\\
\hline
\end{tabular}
\end{center}
\caption{Details of V-CNN I. It has about 3.5M parameters.}
\end{table*}

We use all $60$ orientations of all objects in the training set to train the network. During test time, these $60$ orientations of each object is passed through the network till the first fully connected layer. A max-pooling layer aggregates the activations from all $60$ orientations before sending it to the final fully connected layer, which is our classifier. In other words, we use weight sharing across all $60$ orientations, helping us achieve better classification performance without any increase in the model size.

\subsubsection{Data Augmentation} \label{subsub:DataAug}
The 3D CAD models in both ModelNet40 and ModelNet10 datasets are in the form of polygon mesh, containing coordinates of all the vertices in the mesh and the ID of each node forming a polygon. In order to help the Volumetric CNN learn invariance to small deformations (a lot of objects in a class can be seen as small deformations of other objects in the class), we augment the dataset with meshes whose vertices have been randomly displaced from its original position. We choose this random displacement from a centered Gaussian distribution with a standard deviation of $5$. This standard deviation was chosen after manually inspecting several CAD model files. We use this noisy dataset along with the original dataset to train V-CNN I.

\subsection{V-CNN II}
\begin{figure*}[t] \label{figure:VCNNII}
\begin{center}
	\includegraphics[width=\textwidth]{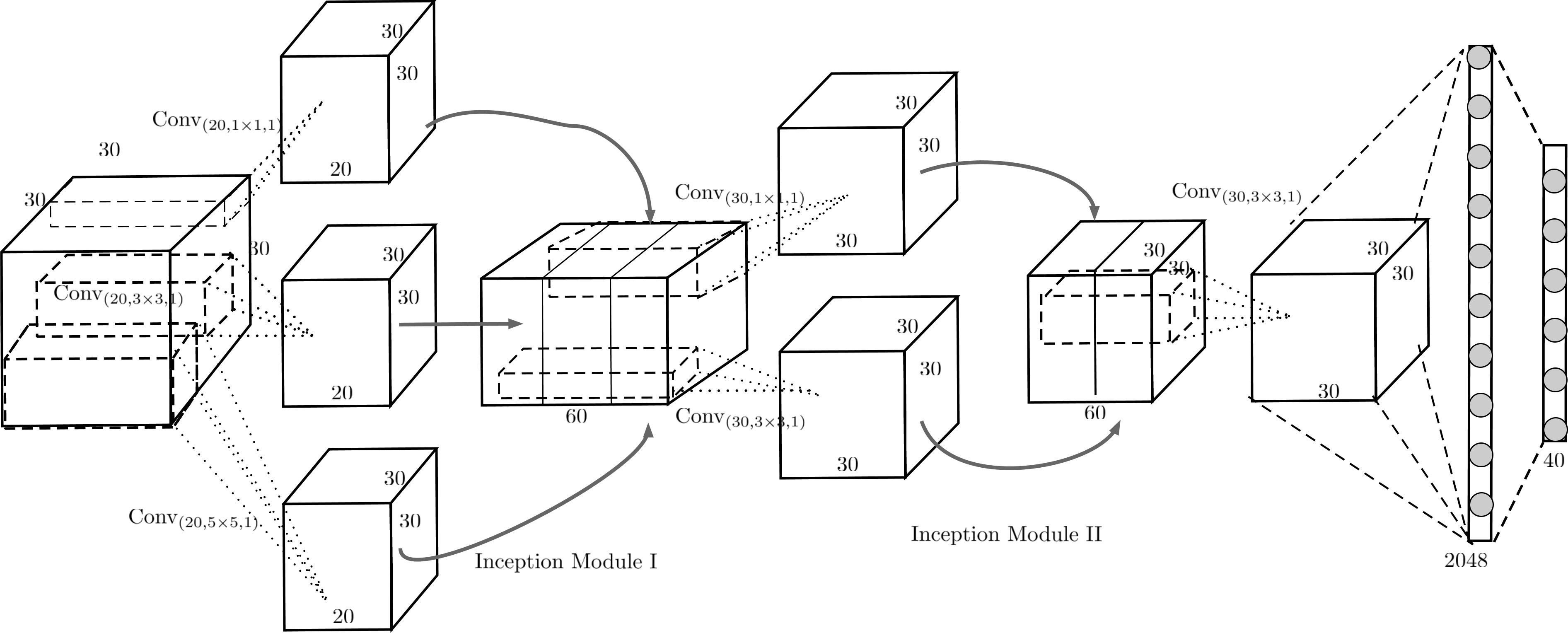}
\end{center}
   \caption{Network architecture of V-CNN II. It has $2$ inception modules. The first inception module is a concatenation of three convolution outputs (of kernel sizes $1 \times 1$, $3 \times 3$, $5 \times 5$). The convolutional layers in the first inception module assumes the number of channels to be $30$. The second inception module is a concatenation of two convolution outputs (of kernel sizes $1 \times 1$ and $3 \times 3$). Softmax loss function is used while training the network.}
\label{voxel}
\end{figure*}

Inspired by the inception module in GoogLeNet \cite{DBLP:journals/corr/SzegedyLJSRAEVR14}, which concatenates outputs from kernels of different size to capture features across multiple scales, we use an inception module for volumetric data to concatenate outputs from filters of size $1 \times 1$, $3 \times 3$ and $5 \times 5$ as depicted in figure \ref{figure:VCNNII}. The usage of $1 \times 1$ kernel is based on the Network in Network (NIN) ~\cite{DBLP:journals/corr/LinCY13} idea which abstracts information in the receptive field and encodes a higher representational power without much additional computational cost. These voxels would otherwise have been processed directly by a convolution kernel. While the inception module uses $1 \times 1$ filter before applying $3 \times 3$ or $5 \times 5$ filters in order to reduce computational complexity, we do no such thing since the model by itself is very small compared to many state of the art 2D image classification models. We use two such inception modules in our network, followed by a convolution layer and two fully connected layers. We use dropout to reduce any over-fitting. As in V-CNN I, a fully connected layer with $40$ neurons is used as the classifier for ModelNet40 dataset and a fully connected layer with $10$ neurons as the classifier for ModelNet10 dataset. Details of this architecture is given in table \ref{table:VCNN II}.

\begin{table*}
\begin{center}
\label{table:VCNN II}
\begin{tabular}{|c|c|c|c|c|}
\hline
Layer type & Filter size/Dropout rate & Stride & Output size & Number of parameters\\
\hline\hline
Convolution & $1 \times 1$ & 1 & $20 \times 30 \times 30$ & 620\\
Convolution & $3 \times 3$ & 1 & $20 \times 30 \times 30$ & 5420\\
Convolution & $5 \times 5$ & 1 & $20 \times 30 \times 30$ & 15020\\
Concat & -- & -- & $60 \times 30 \times 30$ & --\\
ReLU & -- & -- & $60 \times 30 \times 30$ & -- \\
Dropout & 0.2 & -- & $60 \times 30 \times 30$ & --\\
& & & &\\
Convolution & $1 \times 1$ & 1 & $30 \times 30 \times 30$ & 1830\\
Convolution & $3 \times 3$ & 1 & $30 \times 30 \times 30$ & 16230\\
Concat & -- & -- & $60 \times 30 \times 30$ & --\\
ReLU & -- & -- & $60 \times 30 \times 30$ & -- \\
Dropout & 0.3 & -- & $60 \times 30 \times 30$ & --\\
& & & & \\
Convolution & $3 \times 3$ & 1 & $30 \times 30 \times 30$ & 16230\\
ReLU & -- & -- & $30 \times 30 \times 30$ & -- \\
Dropout & $0.5$ & -- & $30 \times 30 \times 30$ & --\\
& & & & \\
Fully Connected & -- & -- & 2048 & 55,298,048\\
Fully Connected & -- & -- & 40 & 81,960\\
\hline
\end{tabular}
\end{center}
\caption{Details of V-CNN II. It has about 55M parameters.}
\end{table*}

Similar to V-CNN I, we use weight sharing across all $60$ orientations. We aggregate activations from all $60$ orientations of the voxelized model from the first fully connected layer using a max-pool.

\subsection{Classification}
All $60$ randomly generated orientations are used to separately train both networks end to end using Caffe \cite{jia2014caffe}. Each of these orientations are treated as different objects while training. The loss function used is Softmax loss, which is a generalization of logistic loss for more than two classes. The weights are updated using Stochastic Gradient Descent with momentum and weight decay. During test time, the $60$ views are passed through the network to obtain features from the first fully connected layer after which, a max-pooling layer pools these features across all $60$ views before sending it to the final fully connected layer which performs the classification.

For ModelNet40, we train both V-CNN I and V-CNN II from scratch. i.e. using random weight initialization. For ModelNet10 dataset we finetune the weights obtained after training the networks on ModelNet40. This method is similar to MV-CNN proposed in \cite{MVCNN} which uses weights from VGG-M network pre-trained on ImageNet. This gives a superior performance than using random weight initialization, demonstrating the power of transferring features learned on a big dataset. This also makes a case for building bigger 3D model repositories, perhaps on the scale of ImageNet.

\section{Multi-View CNN}
Multi-View CNN (MV-CNN), introduced in \cite{MVCNN} aggregates 2D projections of the polygon mesh using a standard VGG-M network. This achieved state of the art classification accuracy as seen on the Princeton ModelNet leader-board. We use AlexNet instead of VGG-M and obtain accuracies far better than our Volumetric CNNs explained in section \ref{section:V-CNN} using only $20$ views, unlike in \cite{MVCNN} where for each of the $20$ views, the camera is rotated $0$, $90$, $180$ and $270$ degrees along the axis passing through the camera into the centroid of the object, giving $80$ views per object. Recent work on Active Multi-View Recognition \cite{DBLP:journals/corr/JohnsLD16} predicts the Next Best View (NBV) which is most likely to give the highest extra information about the object, needing a smaller number of image sequences during test-time to predict the model class. Similar to our V-CNN, we use weight sharing across all $20$ views to keep the model size intact, while achieving a superior performance.

\begin{figure}[t] 
\begin{center}
	\includegraphics[scale=0.45]{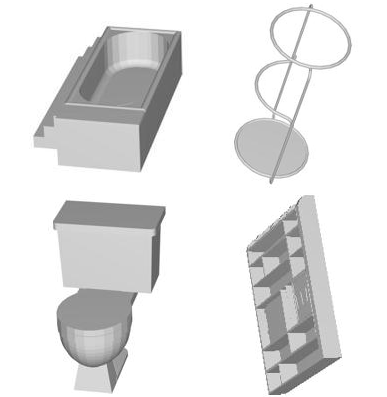}
\end{center}
   \caption{Sample 2D projections of 3D CAD models from ModelNet40. Top-left: bathtub, top-right: stool, bottom-left: toilet, bottom-right: wardrobe. Phong reflection method was used to obtain all 2D projections \cite{Phong:1975:ICG}}
\label{voxel}
\end{figure}

\subsection{Classification}
We render multiple 2D projections of a polygon mesh using cameras placed on the $20$ corners of an icosahedron and use all $20$ views for training and testing. The projections obtained are grey-scale images since the original polygon mesh dataset does not have any color information in it. We use an AlexNet model pre-trained on ImageNet before fine-tuning it for 2D projections. Since the 2D projections we obtain are gray-scale, we replicate pixel values so that we obtain a 3-channel image which can be used in AlexNet. We use the Softmax Loss function for training, which is a generalization of logistic loss for multiple target classes. While training the network, we used a base learning rate of $0.001$ and a momentum value of $0.9$.

One of the reasons for superior performance of Multi-View CNN compared to Volumetric CNN is that we use transfer learning to warm start the weights learned from ImageNet. We know that weights learned in the initial layers of a neural network on one dataset generalize well to a different dataset \cite{DBLP:journals/corr/YosinskiCBL14}. This is not true for Volumetric CNNs due to the absence of very large datasets for 3D models. Since earlier layers in a CNN learn very basic features (for example, the first layer learns features similar to Gabor filter) \cite{DBLP:journals/corr/YosinskiCBL14} which are not specific to the dataset, we only fine-tune later fully connected layers, starting from $\it{fc6}$ onward in AlexNet while keeping the remaining layers frozen. While finetuning, we consider all $20$ views to be distinct images and train AlexNet end to end. While testing, we pass all the $20$ views of a 3D object through to the second fully connected layer, $\it{fc7}$. At this point a max-pool layer finds the maxima for each neuron across all 20 views. This aggregate of the $20$ views are sent to \textit{fc8} for classification, similar to what was done in \cite{MVCNN}.

\section{Experiments} \label{section:Experiments}
We first compare the performance of individual networks separately and then combine two or more networks to improve the classification accuracy. We do this for both ModelNet10 and ModelNet40 datasets. The summary of our findings can be found in table \ref{table:acc_comparision}. It can be seen that V-CNN I and V-CNN II perform similarly. However, it is important to note that they learn slightly different features and therefore can be combined to produce a slightly better classifier as can be seen from the table. Multi View CNN based on AlexNet performs significantly better than Volumetric CNN. 

One of the reasons for this difference is due to gains from finetuning on a network pre-trained on ImageNet. For ModelNet40 dataset, can be seen that there is a gain of $1.1$ percent when finetuned on just the last three fully connected layers of AlexNet. A gain of about $1.6$ percent can be seen when finetuned MV-CNN based on AlexNet is combined with V-CNN I. This means that there is significant non-overlap in the strengths of the two individual networks, allowing us to obtain a better classification accuracy.

Finally, we present FusionNet, an aggregation of V-CNN I trained on $60$ views, V-CNN I trained on 60 views (with augmented data), V-CNN II trained on $60$ views and MV-CNN based on AlexNet and pretrained on ImageNet. It is the best performing combination that we've obtained so far for both ModelNet10 and ModelNet40.

\subsection{FusionNet}
An ensemble of networks have been shown to boost the classification performance for 2D RGB image data. Unsurprisingly, we find it to be true for CAD models as well. However, while most ensemble methods have a single data representations for input, we use multiple data representations to achieve an enhanced classification capability. While 2D projections capture locally spatial correlations as features, voxels can be used to capture long range spatial correlations which will also contribute to discriminative power of a classifier. 

We combine these networks (trained on both voxel and pixel representations) after the final fully connected layer which outputs class scores. A linear combination of class scores is taken with the weights obtained through cross-validation. The class corresponding to the highest score is declared the predicted class.

\begin{figure}[t] 
\begin{center}
	\includegraphics[scale=0.27]{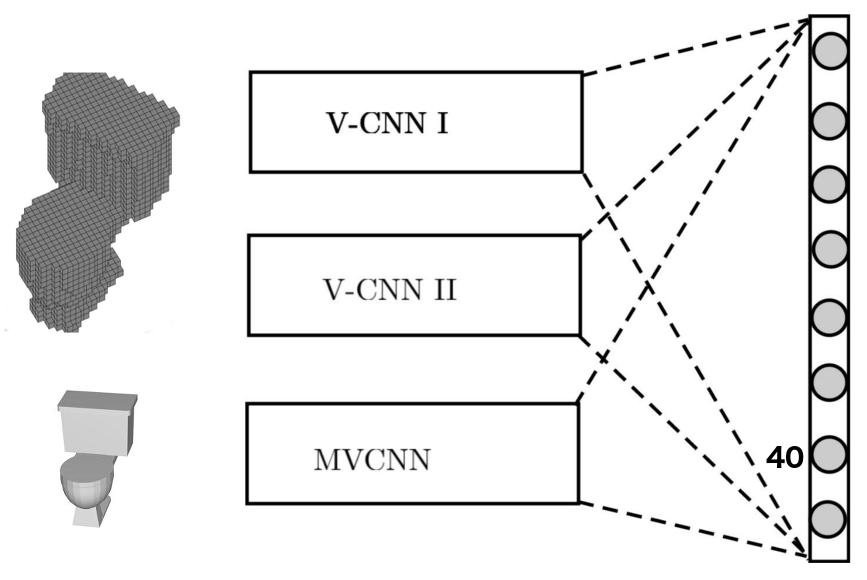}
\end{center}
   \caption{FusionNet is a fusion of three different networks: V-CNN I, V-CNN II and MV-CNN (which is based on AlexNet and pre-trained on ImageNet). The three networks fuse at the scores layers where a linear combination of scores is taken before finding the class prediction. Voxelized CAD models are used for the first two networks and 2D projections are used for latter network}
\label{voxel}
\end{figure}

\begin{table*} \label{table:acc_comparision}
\begin{center}
\label{CNN_tuning}
\begin{tabular}{|c|c|c|c|}
\hline
Network & Number of Views Used & Accuracy (ModelNet10) & Accuracy (ModelNet40)\\
\hline\hline
Volumetric CNN (V-CNN) 1 & 60 & 91.48 & 82.41\\
V-CNN I* & 60 & -- & 80.63\\
V-CNN II & 60 & 90.22 & 82.11\\
V-CNN II + V-CNN II & 60 & 90.32 & 83.31\\
V-CNN I + V-CNN II & 60 & 91.95 & 83.78\\
AlexNet (random) MV-CNN & 20 & -- & 85.82\\
AlexNet (FT) MV-CNN & 20 & 92.69 & 86.92\\
AlexNet (FT) MV-CNN + V-CNN I & 20, 60 & 93.04 & 88.50\\
\bf{FusionNet} & \bf{20, 60} & \bf{93.11} & \bf{90.80}\\
\hline
\end{tabular}
\end{center}
\caption{Classification accuracy for individual models. FT = Fine Tuning. * = Data augmentation with Gaussian noise. random = Random initialization of all network weights. FusionNet achieves the best performance.}
\end{table*}

\section{Conclusions and Future Work}
In our work, we have shown the importance of two different representations of 3D data which can be used in conjunction for classifying 3D CAD models. In our experiments, we see that networks based on the two representations individually perform well on partially non-overlapping set of objects which we believe stems from representation dependent features learned from the two representations. We show that combining the two networks yields a significantly better performance than each of the individual networks and discuss the probable reasons for such a significant boost in performance. 

While guessing the Next Best View (NBV) based on the current view has shown promise on 2D projections based method \cite{DBLP:journals/corr/JohnsLD16}, we believe that a similar method can be applied to volumetric representation, requiring far fewer orientations during test time in order to achieve similar classification accuracy. Other representations of 3D images like signed/unsigned distance fields can be explored to see if they provide significant classification enhancement.

\subsubsection*{Acknowledgments}
We gratefully acknowledge Matroid Inc. for funding this project and for providing compute time on AWS GPU instances. We thank Ameya Joshi and Abhilash Sunder Raj for their help in designing V-CNN I. 

{\small
\bibliographystyle{ieee}
\bibliography{egbib}
}

\end{document}